\documentclass{article}

\usepackage{microtype}
\usepackage{graphicx}
\usepackage{subcaption}
\usepackage{booktabs} 

\usepackage{hyperref}

\usepackage{amsmath,amsbsy}
\newcommand*{\fatplus}{\ensuremath{\boldsymbol{\pmb{+}}}}

\usepackage{xspace}
\newcommand*{\SQCross}{\ensuremath{\boldsymbol{\pmb{\Join}}}\xspace}
\newcommand*{\SQPlus}{\fatplus\xspace}
\newcommand*{\SQC}{\ensuremath{\mathrm{\mathbf{C}}}\xspace}
\newcommand*{\SQG}{\ensuremath{\mathrm{\mathbf{G}}}\xspace}
\newcommand*{\SQR}{\ensuremath{\mathrm{\mathbf{R}}}\xspace}
\newcommand*{\SQS}{\ensuremath{\mathrm{\mathbf{S}}}\xspace}
\newcommand*{\SQU}{\ensuremath{\mathrm{\mathbf{U}}}\xspace}
\newcommand*{\SQA}{\ensuremath{\mathrm{\mathbf{A}}}\xspace}
\newcommand*{\SQF}{\ensuremath{\mathrm{\mathbf{F}}}\xspace}
\newcommand*{\SQPrepare}{\ensuremath{\mathrm{\mathbf{G^1}}}\xspace}

\usepackage{amssymb}

\usepackage{latexsym}

\usepackage[accepted]{mlsys2025}

\mlsystitlerunning{Span Queries}

\newtheorem{definition}{Definition}

\begin{document}

\twocolumn[
\mlsystitle{Using Span Queries to Optimize for Cache and Attention Locality}



\mlsyssetsymbol{equal}{*}

\begin{mlsysauthorlist}
\mlsysauthor{Paul Castro}{ibmny}
\mlsysauthor{Nick Mitchell}{ibmny}
\mlsysauthor{Nathan Ordonez}{ibmzh}
\mlsysauthor{Thomas Parnell}{ibmzh}
\mlsysauthor{Mudhakar Srivatsa}{ibmny}
\mlsysauthor{Antoni Viros i Martin}{ibmma}
\end{mlsysauthorlist}

\mlsysaffiliation{ibmny}{IBM Research, New York, USA}
\mlsysaffiliation{ibmma}{IBM Research, Massachusetts, USA}
\mlsysaffiliation{ibmzh}{IBM Research, Zurich, Switzerland}

\mlsyscorrespondingauthor{Nick Mitchell}{nickm@us.ibm.com}

\mlsyskeywords{KV Cache, Locality, Inference}

\vskip 0.3in

\begin{abstract}
Clients are evolving beyond chat completion, and now include a variety of innovative inference-time scaling and deep reasoning techniques. At the same time, inference servers remain heavily optimized for chat completion. Prior work has shown that large improvements to KV cache hit rate are possible if inference servers evolve towards these non-chat use cases. However, they offer solutions that are also optimized for a single use case, RAG. In this paper, we introduce the \emph{span query} to generalize the interface to the inference server. We demonstrate that chat, RAG, inference-time scaling, and agentic workloads can all be expressed as span queries. We show how the critical distinction that had been assumed by prior work lies in whether the order of the inputs matter --- do they \emph{commute}? In chat, they do not. In RAG, they often do. This paper introduces span queries, which are expression trees of inference calls, linked together with commutativity constraints. We describe span query syntax and semantics. We show how they can be automatically optimized to improve KV cache locality. We show how a small change to vLLM (affecting only 492 lines) can enable high-performance execution of span queries. Using this stack, we demonstrate that span queries can achieve 10-20x reductions in TTFT for two distinct non-chat use cases. Finally, we show that span queries can also be optimized to improve \emph{attention locality}, so as to avoid the so-called lost-in-the-middle problem. We demonstrate that an attention-optimized span query on a 2b parameter model vastly outperforms the accuracy of a stock inference server using an 8b model.




\end{abstract}
]



\printAffiliationsAndNotice{}  

\section{Introduction}
\label{Introduction}

Cache locality is core to the viability of transformer-based large language models~\cite{llmd2025kvcache} ~\cite{li2025surveylargelanguagemodel}. The Key-Value (KV) cache amortizes the quadratic complexity of self-attention~\cite{vaswani2017attention}, reducing ``prefill'' GPU load both within and across requests. In this paper we focus on improving cross-request KV cache locality for emerging 
agentic workloads~\cite{PwC_AIAgentSurvey_2025} and for the ever-evolving and innovative suite of inference time scaling (ITS) patterns~\cite{davis2024networksnetworkscomplexityclass,zhou2025evaluatingjudgesevaluatorsjetts}. 


Inference servers such as vLLM~\cite{10.1145/3600006.3613165} are optimized for \emph{prefix-based} memory reuse. For example, in a chatbot the chat history accumulates linearly and thus, with each turn of a chat, the prefix remains the same. Model servers have grown to embody this reuse pattern. In contrast, agentic and ITS workloads combine the output of multiple separate calls, and make extensive use of a corpus of post-training knowledge --- the so-called retrieval augmented generation or RAG pattern~\cite{liang2024kagboostingllmsprofessional,sarthi2024raptorrecursiveabstractiveprocessing}. Neither one these has a prefix-based pattern of reuse. 

Rather, with these emerging workloads, the order in which input is reused changes from request to request. For example, with RAG the first request may retrieve fragments $F_1$ and $F_2$ and present them in that order. The second request may present $F_2$ and $F_3$, thus the order of $F_2$ has changed. The tricky part is that whether or not order matters, i.e. whether or not it is valid to permute the order of input, depends on the application. There is no blanket rule of thumb, as observed by~\cite{yao2025cacheblendfastlargelanguage} in their CacheBlend work. Implicit in their approach is that, since this constraint is not known, it is better to optimize for the case that order does matter. We take a complementary approach.

The central premise of the paper is this: chat, RAG, ITS, and agentic workloads are special cases of a more general structure, and this more general structure hinges on the expression of whether order matters. The algebraic way to express this is via operator \emph{commutativity}: is $A,B$ the same as $B,A$? In \autoref{sec:span-queries}, we introduce the \emph{Span Query}, a declarative intermediate representation that allows one to express how model calls can be arranged into a hierarchical expression tree that is linked together with commutativity constraints. For example, a span query for RAG might express that the retrieved fragments can be sequenced in any order. A span query for a judge/generator can express that the judge can inspect the generated candidates in any order.

If the span query expresses that order matters, techniques such as CacheBlend can be used. If order does not matter,
we shall show in \autoref{sec:locality-cache} how to automatically optimize span queries in order to improve KV cache locality and thus reduce time-to-first-token (TTFT) and prefill GPU load by as much as 20x --- far exceeding the 3--4x reported by CacheBlend. Prior work on Block Attention~\cite{ma2025blockattention} achieves similar gains, but in a way that is hard-coded for a single use case (RAG). Even on cache miss, we shall show that commutative reuse reduces prefill load by 3x due to attention sparsity. We show how we did this with the help of only 492 lines of code changed in vLLM.

Finally, we share a surprising emergent property of span queries: we can also optimize them for \emph{attention locality}. Smaller models struggle to attend to important information located in the middle of long inputs --- termed the ``lost in the middle'' phenomenon by~\cite{liu-etal-2024-lost}. In \autoref{sec:locality-attention}, we show how span queries can be optimized to improve attention locality, by treating the expression tree like a divide-and-conquer tree --- c.f. map/reduce~\cite{dean2008mapreduce}. We show that, as with KV cache locality, attention locality optimization also hinges on commutativity. We demonstrate that an optimized span query on a 2b model outperforms the accuracy of an 8b model. This was not our original design point, and so it was satisfying to find that commutativity constraints have more general applicability.

\section{A Brief Tour of vLLM}
\label{sec:background-vllm}

This section provides an overview of the interaction of vLLM and its key-value (KV) cache. We concentrate on the chat completion API, which is shared with other inference servers. The chat completion API takes as input a sequence of typed \emph{messages}. The most common message types are \texttt{System}, \texttt{User}, and \texttt{Assistant}; in what follows you may see these designated by their first letter (\SQS, \SQU, \SQA). A system message instructs to the model as to how best to respond, a user message represents a question being posed to the model, and an assistant message represents the output of some prior model interaction.
Models are trained on a language that has been encoded into a lexicon of \emph{tokens}. We thus consider the input to be sequences of typed tokens.

Second, vLLM breaks token sequences into fixed (but configurable) size \emph{blocks}.\footnote{The vLLM literature uses ``page'' while the source and we use ``block'' to mean a fixed-size token sequence. \cite{ma2025blockattention} uses ``block'' to mean a variably-sized  sequence --- to us, a ``span''.} All KV cache operations (insertion, lookup, eviction, etc.) are performed at the block granularity. vLLM does not cache partial blocks across requests. For example, if vLLM is configured to have a block size of 16 tokens (the default setting as of this writing) and a message tokenizes to 17 tokens, the last token might not be cached.


Third, cached output blocks (i.e. those \emph{generated} by the chat completion API call) implicitly depend on the input that generated them. This dependence is a product of the Transformers' architecture. The hash code for a block $B_i$, the $i$th in a sequence, depends on the hash code of block $B_{i-1}$. 
Thus, when provided input blocks $[I_1, \ldots, I_n]$ and observing output blocks $[O_1, \ldots, O_m]$, the KV cache can be considered to be keyed by some hash of $[I_1, \ldots, I_n, O_1, \ldots, O_m]$. Furthermore, when performing a cache lookup for a given sequence $[B_1,\ldots,B_n]$, vLLM stops scanning for cache hits after the first miss; i.e. it only searches for a \emph{prefix} of cached blocks $[B_1,\dots,B_{h \le n}]$.

\begin{figure}
    \centering
    \includegraphics[width=0.75\linewidth]{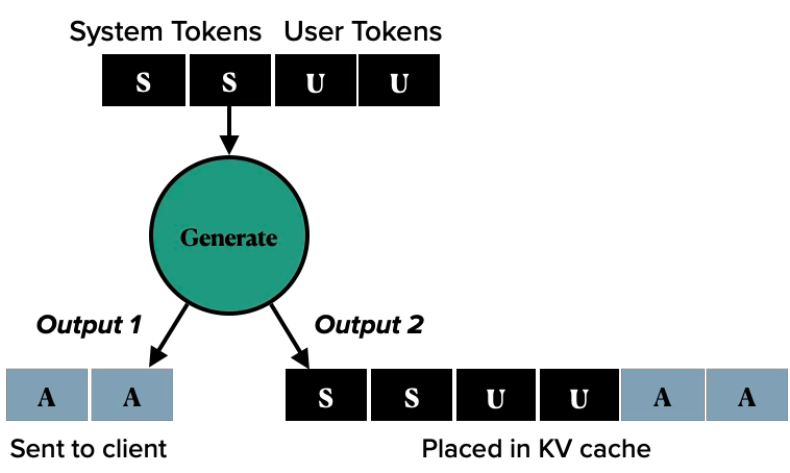}
    \caption{The ``dual output paradox'': the model server emits one thing to the client and something different to KV cache. }
    \label{fig:dual-output-paradox}
\end{figure}

Finally, vLLM's chat completion API can be considered as having two distinct outputs. We term this the \emph{dual output paradox}. One is sent to the client $[O_1,\dots,O_M]$, and the other is inserted into KV cache (as illustrated in \autoref{fig:dual-output-paradox}). In some use cases (\autoref{sec:use-case-chat}), this paradox can be resolved by the client, as it can prefix each request with the history of all prior interactions. For other use cases (\autoref{sec:use-cases}), resolving the paradox requires changes to vLLM such as those introduced in this paper.

\subsection{Use Case: Chat Completion}
\label{sec:use-case-chat}

\begin{figure}
    \centering
    \includegraphics[width=\linewidth]{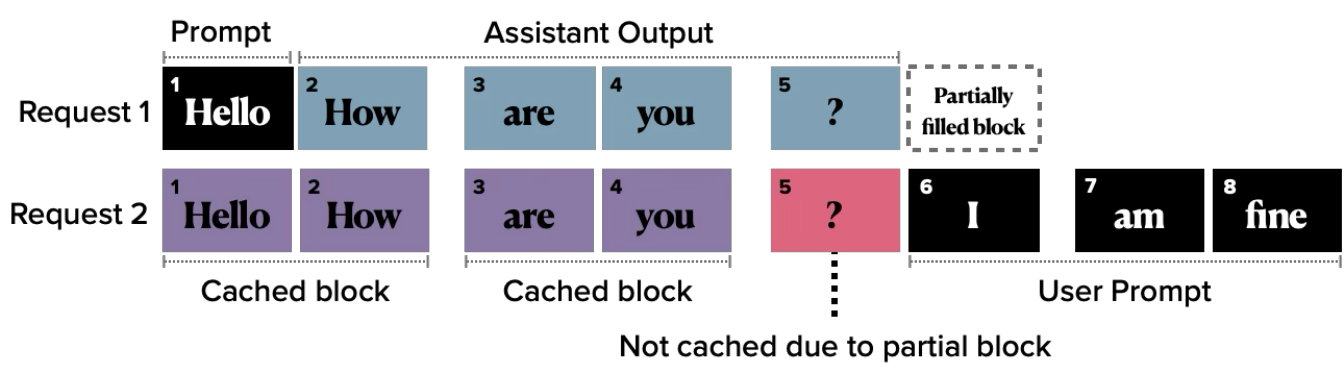}
    \caption{Chat completion use case. Each rectangle is a token, a cache block fits 2 tokens, and a token's sequence position is shown in upper left corner. 80\% hit rate on Request 2 (4 of 5 input tokens are cached), which asymptotes to 100\% as the chat progresses.}
    \label{fig:chat-completion-illustration}
\end{figure}

With that background in mind, we illustrate how well-suited vLLM's architecture is to the chat completion use case.
Consider the example in~\autoref{fig:chat-completion-illustration}. For illustration purposes, the hypothetical block holds 2 tokens. If a user submits ``Hello'' (say 1 token) and the model generates ``How are you?'' (say 4 tokens), then vLLM will cache 2 blocks:  \texttt{[Hello, how]} on the first block and \texttt{[are, you]} on the second block. The final generated token \texttt{?} is not cached for future requests as it lies on a partially filled block.

The chat session continues as the user responds ``I am fine''. The OpenAI chat completion API is stateless. Therefore, the chat client prepends the 5 tokens of chat history (referred to as ``context'' tokens) to the second request, which becomes ``Hello How are you? I am fine''. vLLM can avoid recomputing 4 out of 5 of the context tokens (the fifth, as per above, was not cached due to a partially filled block). 
In our chat completion example cache hit rate approaches 100\% as the chat session progresses. This demonstrates that vLLM is well optimized for the chat completion use case.

With chat completion, blocks maintain their position within the accumulating chat context. For example, for the duration of the chat session from \autoref{fig:chat-completion-illustration} ``Hello'' will always appear first, ``How'' always second, and so on. Chat completion has an append-only pattern of updates. In contrast, while non-chat use cases also have reuse across requests, the position of reused blocks changes from request to request. 
\section{Non-Chat Use Cases}
\label{sec:use-cases}

Next, we introduce the two non-chat use cases: retrieval-augmented generation (RAG) and ``nested generation''. 


\subsection{Use Case: Retrieval-augmented Generation}
\label{sec:use-case-rag}

\begin{figure}
    \centering
    \includegraphics[width=\linewidth]{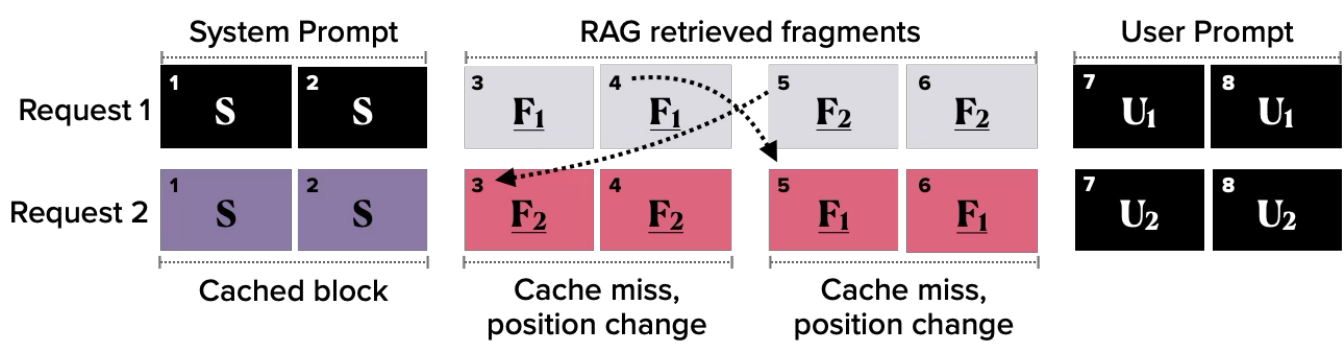}
    \caption{RAG use case.
    33\% hit rate on Request 2 (2 of 6 input tokens cached), which asymptotes to 0\% as $F_1,F_2$ grow.}
    \label{fig:rag-illustration}
\end{figure}

Retrieval-augmented generation (RAG) is a technique for supplementing a chat request with knowledge from an online corpus~\cite{liang2024kagboostingllmsprofessional,sarthi2024raptorrecursiveabstractiveprocessing}. Chat clients that support RAG operate in two phases: indexing and retrieval. \autoref{fig:rag-illustration} illustrates a RAG use case. 

First the corpus is \emph{indexed}. A simple indexing strategy fragments the corpus into contiguous segments and then \emph{embeds} each fragment in a vector space.
In \autoref{fig:rag-illustration}, the corpus consists of embedded fragments $F_1, F_2, \ldots$.

Second, in response to a user message, the client embeds it in the vector space and then \emph{retrieves} a list of nearby vectors. The closest vectors are de-embedded back to text and sent, along with the user's message, to the chat completion server.
In \autoref{fig:rag-illustration}, the first query retrieves fragments in the order $[F_1, F_2]$ and the second query retrieves fragments in the reverse order. Even though there is reuse of the fragments across the requests, vLLM will not consider them to be cache hits because the prefix of $F_2$ in the second request differs from that in the first request. For this reason, as the length of the retrieved fragments grows, RAG use cases will asymptote to a 0\% cache hit rate.

Recent work introduced \emph{block attention}~\cite{ma2025blockattention} with the goal of improving KV cache locality for RAG. We generalize that work in this paper, implementing it within the constraints of the production-grade vLLM model server, and to provide a generalized API that can be automatically optimized to support new use cases.

\subsection{Use Case: Nested Generation}
\label{sec:use-case-nested-generation}

RAG is a proven way to improve output quality without expensive re-training (so-called test-time or inference-time scaling). A complementary inference scaling approach is to wire together model calls into workflows. This allows each model call in a flow to focus on a smaller part of the larger problem. The emerging deep research systems and agentic workflows are examples of this strategy.
Whereas chat completion has an additive and append-only pattern of token accumulation, these use cases have a \emph{nested} structure.


\begin{definition}[Nested Generation]
\label{def:nested-generation}
 A set of model calls $G$ is a \emph{nested generation} if, for any calls $g_1, g_2 \in G$, only the output of $g_2$ (and none of the input to $g_2$) is fed as input to $g_1$. That is, if they nest like conventional function calls.
\end{definition}

\begin{figure}[ht]
    \centering
    \includegraphics[width=0.75\linewidth]{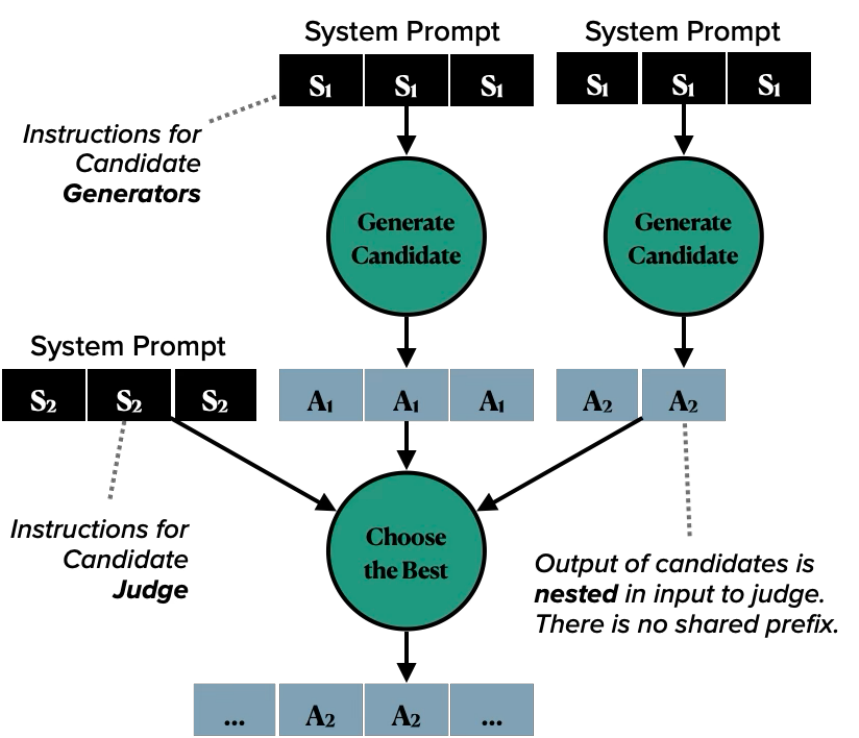}
    \caption{An example of nested generation: the \emph{judge-generator} inference-time scaling strategy.
    }
    \label{fig:judge-generator-flow}
\end{figure}

Consider~\autoref{fig:judge-generator-flow}. This diagram illustrates a common example of nested generation: the \emph{judge-generator} inference-time scaling technique~\cite{davis2024networksnetworkscomplexityclass,zhou2025evaluatingjudgesevaluatorsjetts,snell2025scaling}. Say we want to generate a high quality job application email. Using this technique one first generates a set of candidate emails, instructing the model (via system and user messages) what its job is, and on the specifics of the email (your name, qualifications, etc.). Next, the generated email candidates (the output of the ``inner'' model calls) is passed to the ``outer'' model call. The instructions on what constitutes a good email given to the outer generation will differ from those given to the inner generates. Thus, while there is reuse, from inner to outer model calls, there is no common prefix as we had in the case of chat completion.

\begin{figure}[ht]
    \centering
    \includegraphics[width=\linewidth]{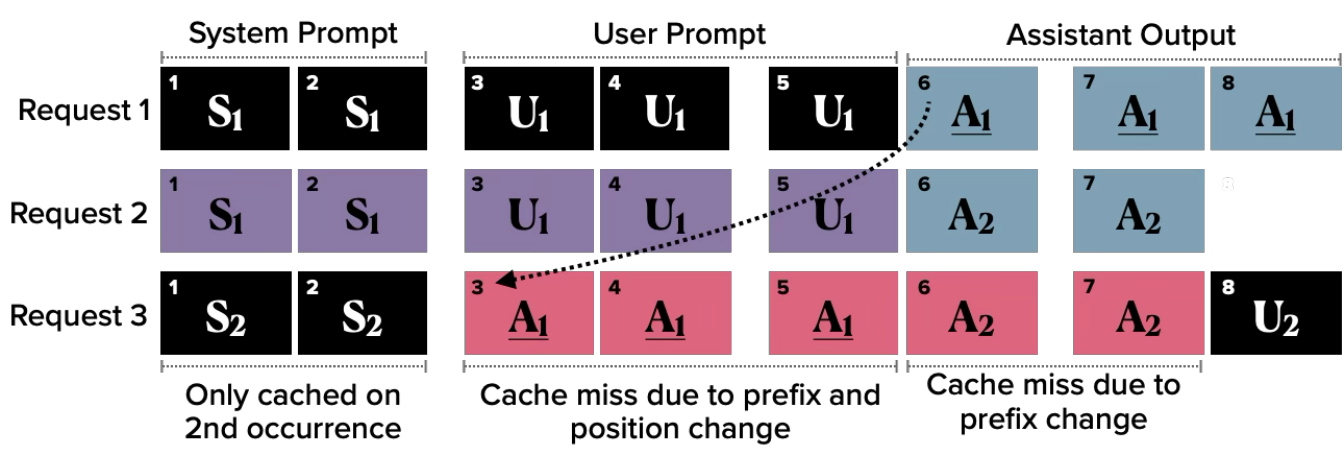}
    \caption{Nested generation use case, following on from \autoref{fig:rag-illustration}. 29\% hit rate on Request 3 (2 of 7 input tokens cached), which asymptotes to 0\% as assistant output grows.}
    \label{fig:nestedgen-illustration}
\end{figure}

\autoref{fig:nestedgen-illustration} illustrates this situation of reuse without a common prefix. In Request 1, a candidate email is generated using system prompt $S_1$ and instructions $U_1$. This produces the candidate email $A_1$ (A for Assistant output). In Request 2, a second email candidate is generated. Here, we get cache hits on $S_1$ and $U_1$, because there is a common prefix. The same is not true for Request 3 which acts as the judge. Even though the tokens from $A_1$ and $A_2$ reside, in some form, in KV cache, they are useless when provided as input to Request 3. The judge has the instructions $S_2 \not = S_1$ as a prefix. As a result, cache hit rate approaches $0\%$ as the lengths of the generated candidates $A_1,A_2$ grow.

\section{The Span Query}
\label{sec:span-queries}

We introduce a declarative intermediate representation, the \emph{span query}.
In this section, we define the syntax and semantics of span queries, provide a graphical representation that helps us to reason about queries, and show how this purely declarative form can be transformed and optimized for high-throughput execution.

\subsection{Query Syntax}

Our goal is to identify a declarative form that expresses our use cases as special cases of a more general structure. To that end, we start with inspiration from the observation made in \autoref{def:nested-generation}: non-chat use cases seem to have the rather conventional structure of a function call graph.  We thus define the span query as a declarative \emph{expression tree} over a set of operators. An expression tree is simply the parse tree of some client-facing library or language. 
We defer the topic of language design to future work. 

\begin{definition}[Span Query]
    A \emph{span query} is an expression tree over the operators $\SQC, \SQR, \SQF, \SQPlus, \SQCross, \SQS, \SQA, \SQU, \SQG$. Consult \autoref{tab:operators} for definitions of the operators.
\end{definition}

\begin{table}
    \caption{A span query is an expression tree over these operators. 
    Desugaring of operators is discussed in \autoref{sec:query-high-level-optimization}. 
    }
    \label{tab:operators}
    \begin{subtable}{0.465\linewidth}
    \begin{center}\begin{small}\begin{sc}
    \caption{Syntactic sugar}
    \begin{tabular}{cl}\toprule
        $\SQC$ & Chat completion \\
        $\SQR$ & corpus Retrieval \\
        $\SQF$ & corpus Fragment \\
        \bottomrule
        \end{tabular}
        \end{sc}\end{small}\end{center}
        \end{subtable}
\quad
    \begin{subtable}{0.46675\linewidth}
    \begin{center}\begin{small}\begin{sc}
    \caption{Message joins}
    \begin{tabular}{cl}\toprule
        $\SQPlus$ & commutative \\
        $\SQCross$ & non-commutative \\
        \bottomrule
        \end{tabular}
        \end{sc}\end{small}\end{center}
    \end{subtable}
\quad
    \begin{subtable}{\linewidth}
    \smallskip
    \begin{center}\begin{small}\begin{sc}
    \caption{Core operators}
    \begin{tabular}{ll}\toprule
        \SQS,\SQA,\SQU & System, User, Assistant messages \\
        $\SQG$ & Generate new tokens \\
        \bottomrule
        \end{tabular}
        \end{sc}\end{small}\end{center}
    \end{subtable}
    \vskip -0.1in
\end{table}

We show in \autoref{sec:query-high-level-optimization} that \SQC, \SQR, \SQF, and \SQPrepare can be desugared and expressed in terms of the core operators. Before we get there, it is helpful to walk through our use cases with a visualization. The visuals demonstrate how each use case is an instance of this general span query structure.



\subsection{Query Visualization}
\label{sec:query-visualization}

We introduce a visualization of span queries. The visuals presented in this paper focus on the relations between the nodes (the edges) rather than node attributes. For example, an \SQS node represents a system message, but the visualization does not show the content of this message. An \SQR node represents retrieval from a corpus, but the visualization does not identify the location of the corpus.
The attributes are nonetheless important, and will come into play when we serialize span queries in \autoref{sec:query-tokenization}.

\begin{figure}
    \centering
    \begin{subfigure}{0.625\linewidth}
        \centering
        \includegraphics[width=0.9\linewidth]{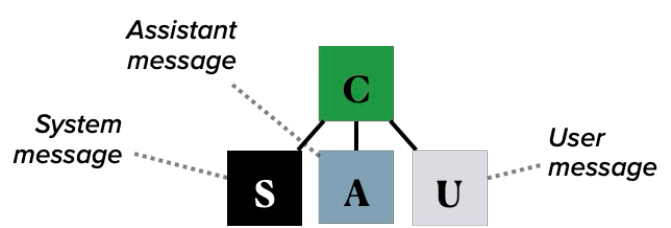}
        \caption{Chat completion}
        \label{fig:chat-completion-query-as-written}
    \end{subfigure}
    \quad
    \begin{subfigure}{0.24\linewidth}
        \smallskip
        \centering
        \includegraphics[width=0.9\linewidth]{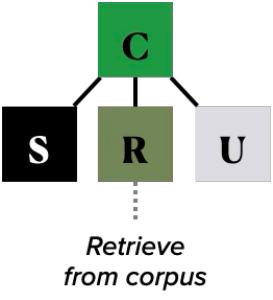}
        \caption{RAG}
        \label{fig:rag-query-as-written}
    \end{subfigure}
    \quad
    \begin{subfigure}{0.465\linewidth}
        \smallskip
        \centering
        \includegraphics[width=0.9\linewidth]{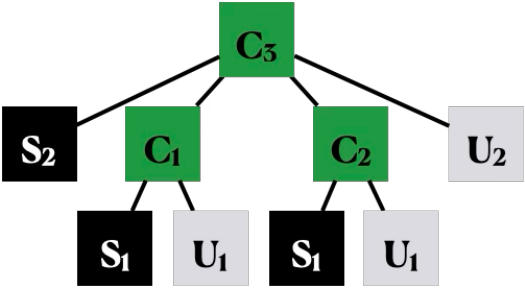}
        \caption{Nested generation}
        \label{fig:nested-generation-query-as-written-naive-c}
    \end{subfigure}
    \quad
    \begin{subfigure}{0.42\linewidth}
        \smallskip
        \centering
        \includegraphics[width=0.8\linewidth]{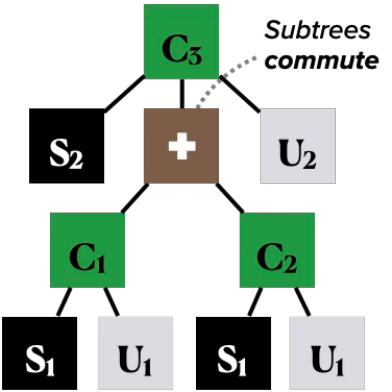}
        \caption{With commutativity hint}
    \label{fig:nested-gen-query-with-plus-c}
    \end{subfigure}
    \caption{Example span queries.}
    \label{fig:span-queries-as-written}
\end{figure}

Consider the visualization of a chat completion span query shown in \autoref{fig:chat-completion-query-as-written}. 
At the root of this span query is the \SQC operator. This operator represents chat completion with input provided by \emph{joining} its children together. Conventionally speaking, the input to \SQC is the \emph{prompt} to the chat completion. In this case, the \SQC operator is fed three typed messages: a system message \SQS, the output of the prior turn of chat \SQA, and the user's question \SQU.

The chat query as shown does not explicitly govern how those three children should be joined together when presented to the model server. For chat completion it may seem obvious, i.e. that the three children should be concatenated in the order given. For chat completion, order matters. For example, one should expect lower quality output if the system message \SQS were placed at the end, or if \SQU came before \SQA. The order of the messages matters for causal attention~\cite{vaswani2017attention}, because later messages can attend to earlier messages but not vice versa. There is an asymmetry, and a concomitant lack of commutativity. 



\autoref{fig:rag-query-as-written} visualizes a RAG query. The structure is nearly identical to the chat completion query from \autoref{fig:chat-completion-query-as-written}, with \SQR replacing \SQA. The \SQR represents retrieval from a corpus. Implicit in retrieval \SQR is a second ordering constraint: how should the retrieved fragments be joined together when presented as input to the \SQC? It seems reasonable to infer a default join semantics from the \SQR node, that order does not matter. We can automatically desugar them in a way that makes ordering constraint explicit.

\autoref{fig:nested-generation-query-as-written-naive-c} visualizes a nested generation query. Here again, are several layers of ordering semantics that need to be teased out. For example, if the output of the inner generates can be safely permuted, then it is a fruitful exercise to reuse their KV cache entries. 
Some instances of nested generation may permit permuting the order of the inner generates, others may not. We need more semantic information from the user before optimization is allowed or effective.




\subsection{Query Semantics}
\label{sec:joins}

A span query is declarative, which means query execution is free of side effects. From this, it is clear cut that every sub-tree of a node can be safely dispatched in parallel. The output of every execution is another span query, thus the set of valid span queries is closed under query execution. 

There is more room for interpretation regarding how a query should \emph{join} child sub-trees in order to construct the input to the parent. This structure, of parallel fork but semantically constrained join, has strong analogies to the map-reduce paradigm~\cite{dean2008mapreduce}.



For our purposes, the semantics of joining messages depends largely on  \emph{commutativity} required by the query. Does order matter, $(M_1,M_2)$ versus $(M_2,M_1)$ for that given pair of messages? The answer will vary by use case, hence a span query must allow that choice to be expressed. 


\begin{definition}{Span Query Join Operators}
A span query can express a message ordering constraint with two operators. \SQCross indicates a non-commutative join of the inputs. \SQPlus indicates the opposite, a commutative join.
\end{definition}

For example, \autoref{fig:nested-gen-query-with-plus-c} shows how a user might amend the nested generation span query from \autoref{fig:nested-generation-query-as-written-naive-c} to introduce a commutativity hint \SQPlus. This hint serves to relax the ordering constraints, thus providing more opportunities for the optimizations covered in the remainder of this paper.




\section{Improving KV Cache Locality}
\label{sec:locality-cache}

We now describe how span queries, with help from a small change to vLLM, can be optimized for KV cache locality. This involves four parts: a) a high-level optimizer that rewrites span queries; b) a span query tokenizer; c) a low-level optimizer that rewrites token sequences; d) the aforementioned small updates to vLLM.
All told, the changes described in this section have a remarkably small impact on vLLM. We added or modified a total of 492 lines of Python source code out of 260,000 across all of vLLM. \autoref{tab:vllm-source-breakdown} breaks down our changes by source file. 

\begin{table}
\caption{Lines of Code (LoC) changed or added to vLLM.
}
\label{tab:vllm-source-breakdown}
\vskip 0.1in
\begin{center}
\begin{small}
\begin{sc}
\begin{tabular}{ll}\toprule
vLLM Source File & LoC \\
\midrule
\texttt{core/block\_pool.py} & 68 \\
\texttt{core/kv\_cache\_manager.py} & 89 \\
\texttt{core/kv\_cache\_utils.py} & 73 \\
\texttt{core/sched/output.py} & 4 \\
\texttt{core/sched/scheduler.py} & 19 \\
\texttt{core/single\_type\_kv\_cache\_manager.py} & 4 \\
\texttt{worker/gpu\_model\_runner.py} & 232 \\
\midrule
Total & 492 \\
\bottomrule
     \end{tabular}
\end{sc}
\end{small}
\end{center}
\vskip -0.1in
\end{table}

\subsection{High-level Cache Locality Optimizations}
\label{sec:query-high-level-optimization}

We break query optimization for KV cache locality into two layers, the first of which is covered here. This high-level optimizer operates on  expression trees. It transforms them to be more explicit in their semantics, to resolve dual output paradoxes, and to simplify them to remove redundancies.

\begin{figure}
    \centering

    \begin{subfigure}{0.253\linewidth}
        \centering
        \includegraphics[width=\linewidth]{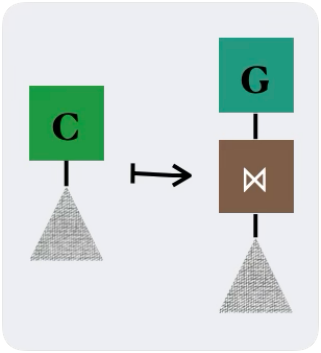}
        \caption{\SQC desugaring}
        \label{fig:hlo-rule-sqc-desugar}
    \end{subfigure}
    \qquad
    \begin{subfigure}{0.325\linewidth}
        \centering
        \includegraphics[width=\linewidth]{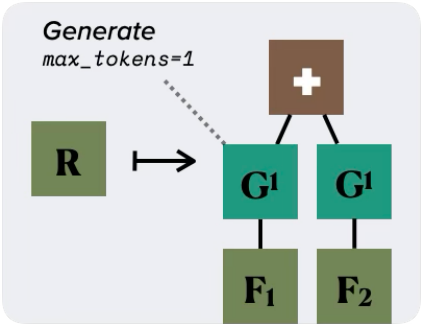}
        \caption{\SQR desugaring}
        \label{fig:hlo-rule-sqr-desugar}
    \end{subfigure}
    \qquad
    \begin{subfigure}{0.397\linewidth}
        \centering
        \includegraphics[width=\linewidth]{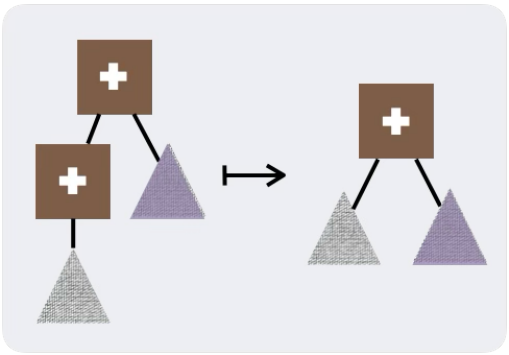}
        \caption{Plus simplification}
        \label{fig:hlo-rule-plus-plus-simplify}
    \end{subfigure}
    \qquad
    \begin{subfigure}{0.325\linewidth}
        \centering
        \includegraphics[width=\linewidth]{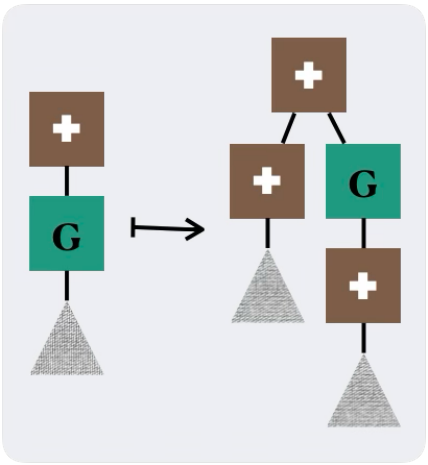}
        \caption{Plus distribution}
        \label{fig:hlo-rule-plus-distribution}
    \end{subfigure}
    
    \caption{Some tree rewriting rules from the high-level optimizer. 
    }
    \label{fig:hlo-rules}
\end{figure}

The high-level optimizer executes a suite of tree transformation in a fixed point iteration. Four such rules are illustrated in \autoref{fig:hlo-rules}, using the visualization from \autoref{sec:query-visualization}. Each illustration shows the left-hand side and right-hand side of a query-to-query transformation. The optimizer looks for matches in a given tree, applies the first match, splices in the replacement sub-tree, and then iterates this process until convergence.

\begin{figure}
    \centering
    \begin{subfigure}{0.325\linewidth}
        \centering
         \includegraphics[width=0.63\linewidth]{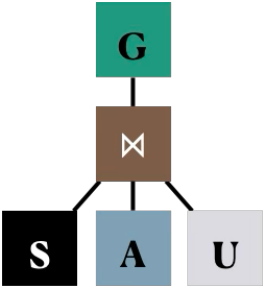}
        \caption{Chat (optimized)}
    \end{subfigure}
    \qquad
    \begin{subfigure}{0.525\linewidth}
        \centering
        \includegraphics[width=0.63\linewidth]{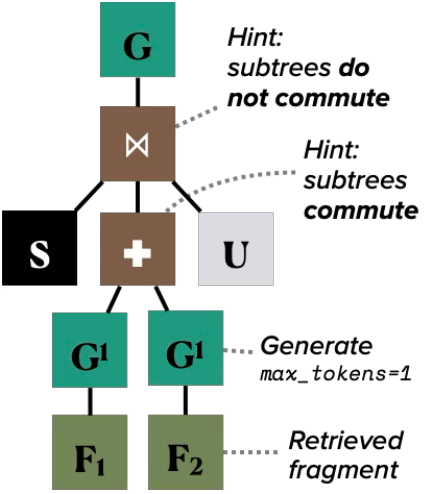}
        \caption{RAG (optimized)}
        \label{fig:rag-query-optimized}
    \end{subfigure}
    \begin{subfigure}{0.55\linewidth}
        \smallskip
        \includegraphics[width=0.88\linewidth]{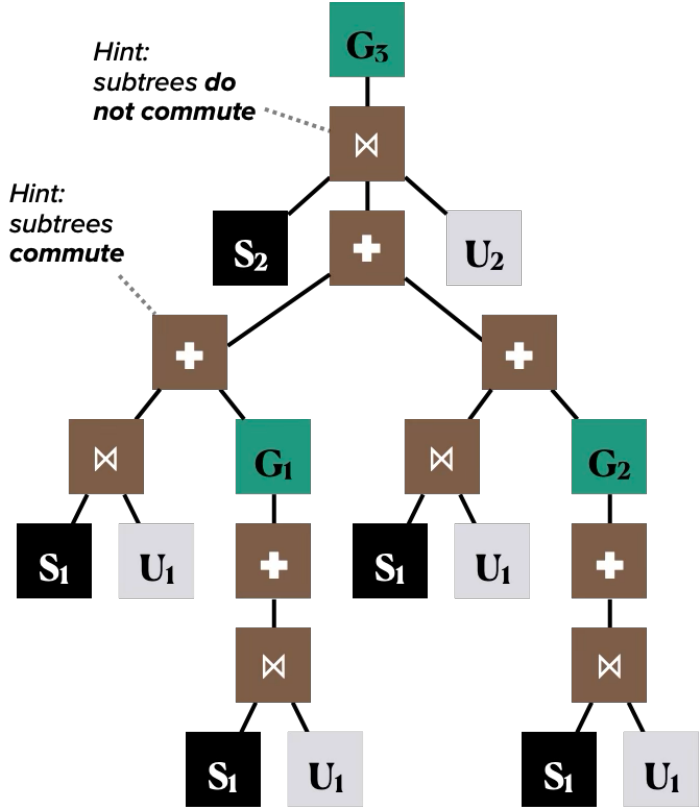}
        \caption{Nested generation (optimized)}
        \label{fig:span-query-nested-gen-optimized}
    \end{subfigure}
    \caption{Optimized span queries. Notice that RAG, after optimization, becomes a special case of nested generation.
    }
    \label{fig:span-query-chat-rag-optimized}
\end{figure}

\autoref{fig:hlo-rule-sqc-desugar} and \autoref{fig:hlo-rule-sqr-desugar} show two of a family of desugaring rules.
\autoref{fig:hlo-rule-sqc-desugar} replaces a \SQC with \SQG (generate new tokens) linked to an explicit \SQCross (non-commutative join of children messages). \autoref{fig:hlo-rule-sqr-desugar} replaces a retrieval node \SQR with an explicit commutative join \SQPlus over the retrieved fragments. The desugaring rule for \SQR relies on \SQPrepare, which represents token generate \SQG but with \texttt{max\_tokens} parameter set to 1. This allows vLLM to \emph{prepare} the fragments prior to their use in the full query.\footnote{There is an asymmetry between cache lookup and insertion. It is straightforward to modify vLLM to avoid biasing at insertion time. It would be a monumental effort to change vLLM's behavior w.r.t. prefilling. Instead, we opt to ``prepare'' the fragments in a way that is independent of context. Span queries can express this.}
\autoref{fig:span-query-chat-rag-optimized} illustrates the optimized forms of the chat completion and RAG queries from earlier.

\autoref{fig:hlo-rule-plus-plus-simplify} shows one of a family of tree simplification rules. This rule leverages the commutativity of \SQPlus to avoid an unnecessary chain of commutativity hints.

\begin{figure}
    \centering
    \begin{subfigure}{0.28\linewidth}
        \centering
        \includegraphics[width=0.58\linewidth]{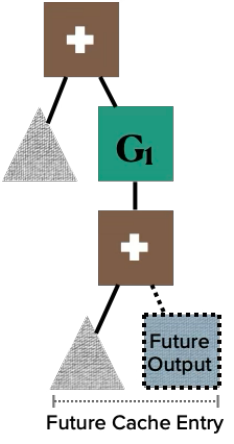}
        \caption{Optimized} 
    \end{subfigure}
    \qquad
    \begin{subfigure}{0.62\linewidth}
        \centering
        \includegraphics[width=0.25\linewidth]{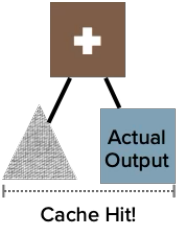}
        \caption{Eventually presented to outer generate}
    \end{subfigure}    
    \caption{Plus distribution resolves the dual output paradox, because it aligns query structure with the behavior of model servers. 
    }
    \label{fig:plus-distribution-explanation}
\end{figure}

\autoref{fig:hlo-rule-plus-distribution} illustrates the ``plus distribution'' rule, which is used to resolve the dual output paradox. 
This rule is valid, because it is valid to distribute the commutative \SQPlus across token generation \SQG. The rule is also effective. Once we have distributed the \SQPlus inside of the \SQG, we ensure that future use of the \emph{output} of the token generation \SQG will be prefixed by the input; this is precisely the constraint underlying the dual output paradox. The effectiveness of this ``ahead of time'' planning is illustrated in \autoref{fig:plus-distribution-explanation}. \autoref{fig:span-query-nested-gen-optimized} illustrates the optimized form of the example query from \autoref{fig:nested-gen-query-with-plus-c}.

\subsection{Query Tokenization}
\label{sec:query-tokenization}

First, we must encode span queries into a form that vLLM can consume. This encoding step allows clients to operate on queries of arbitrary complexity, while allowing vLLM to do its job with minimal changes. Since the core of vLLM operates on token sequences, we must ``parenthesize'' a query. The parentheses will denote the boundaries of each sub-tree in the serialized form  --- c.f.~\cite{merth2024superpositionpromptingimprovingaccelerating}.

We considered but ultimately decided against several approaches. Prior work~\cite{ma2025blockattention} assumes that a list of spans is given as input. To support span queries of arbitrary nesting, we need something more general than that. We could introduce a separate reverse index data structure that maps sequence index to metadata about the sub-tree in which that token resides. It also may be possible to dig more deeply into the vLLM logic so that it accepts trees rather than sequences. We defer these approaches to future work, as they will necessitate large changes to vLLM. 

Our chosen span query encoding produces minimal impact on vLLM. The general approach should not be surprising: employ special tokens to represent the parenthesization. \autoref{tab:low-level-optimization-special-tokens} defines the special tokens. In practice, we can combine these so as to reduce the number of special tokens to two; e.g. \textbf{(} can be reused in place of \texttt{()}, and whitespace is usually a safe pad token. 

\begin{definition}[Span]
In a token sequence $T=[T_1,\ldots,T_n]$, a \emph{span} is a subsequence of $S \subseteq T$ such that $S=[T_i,\ldots,T_j]$ and $T_i$ and $T_j$ are parentheses tokens.
\end{definition}

\begin{table}
    \centering
    \begin{tabular}{ll}
    \toprule
    Tokens & Meaning \\
    \midrule
    $\Box$ & Padding to achieve block alignment \\
    (      & Start of independent sub-tree \\
    )(     & Boundary between independent siblings \\
    )$n$   & End independent sub-tree of length $n$ tokens \\
         \bottomrule
    \end{tabular}
    \caption{Special tokens used by query tokenization.}
    \label{tab:low-level-optimization-special-tokens}
\end{table}

\begin{figure}
    \centering
    \includegraphics[width=\linewidth]{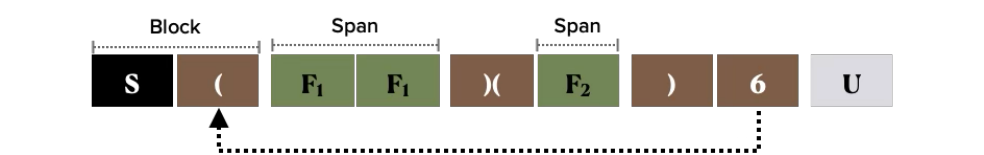}
    \caption{Tokenized RAG span query from \autoref{fig:rag-query-optimized}. Block size of 2 tokens. $S_1, U_1, F_2$ are each messages of 1 token and $F_1$ occupies 2 tokens. The \textbf{6} points to the start of the \texttt{plus} sub-tree. 
    }
    \label{fig:query-tokenization-example}
\end{figure}

To further minimize changes to vLLM, we use a minor variant of this special token approach. We append to the end token \texttt{)} a number that points to the sequence position at which that serialized sub-tree begins. This design complicates the span query tokenization logic, but allows us to avoid some complexity in vLLM proper --- code to maintain the parsing stack necessary to support nested parenthesization. This strategy is illustrated in \autoref{fig:query-tokenization-example}.

Some caution is needed when using special tokens. There is no free lunch. Novel special tokens require model fine tuning. Overloading the meaning of existing tokens may reduce accuracy. We take the latter approach. 

\subsection{Query Low-level Optimization}
\label{sec:query-low-level-optimization}


In order to maximize KV cache locality, reduce overheads, and simplify our vLLM footprint, a serialized span query may need to be rewritten. In this paper, we address two concerns: block alignment and coping with vLLM's propensity to avoid caching trailing partial blocks.

\begin{figure}
    \centering
    \includegraphics[width=\linewidth]{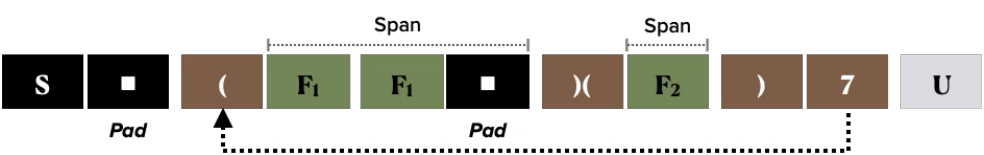}
    \caption{RAG span query after low-level optimizations are applied to the tokenized span query in \autoref{fig:query-tokenization-example}. The \textbf{7} points to the start of the \texttt{plus} sub-tree. \autoref{tab:low-level-optimization-special-tokens} defines the special tokens.}
    \label{fig:rag-query-after-low-level-optimization}
\end{figure}

\subsubsection{Block Alignment}
We rewrite a serialized span query to ensure that the special tokens in a sequence are aligned to block boundaries. We do this by employing the padding special token from \autoref{tab:low-level-optimization-special-tokens}. \autoref{fig:rag-query-after-low-level-optimization} illustrates how padding the serialized query from \autoref{fig:query-tokenization-example} can be used to ensure block boundary alignment.

By enforcing block alignment, the upcoming logic in \autoref{sec:vllm-scheduler-changes} can operate under that assumption. Rather than scanning every token in a block for special tokens, we can limit this search to the first token in every block. This also ensures that all special tokens in cached spans remain in KV cache, as they will only be part of partially filled blocks if the length of a span is smaller than the length of a block. 

\subsubsection{Trailing Partial Blocks}

We may need to rewrite a serialized span query for the nested generation use case. If the output of an inner generate does not fully fill the last block, then there will be a disparity between what is sent back to the client and what is stored in KV cache --- recall the dual output paradox from \autoref{fig:dual-output-paradox}. 

\begin{figure}
    \centering
    \begin{subfigure}{\linewidth}
        \centering
        \includegraphics[width=\linewidth]{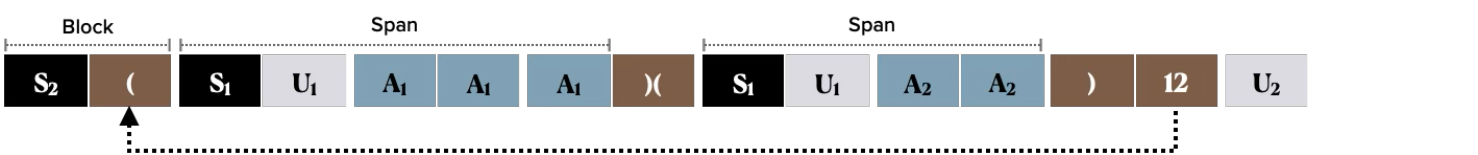}
        \caption{After query tokenization}
    \end{subfigure}
    \begin{subfigure}{\linewidth}
        \centering
        \includegraphics[width=\linewidth]{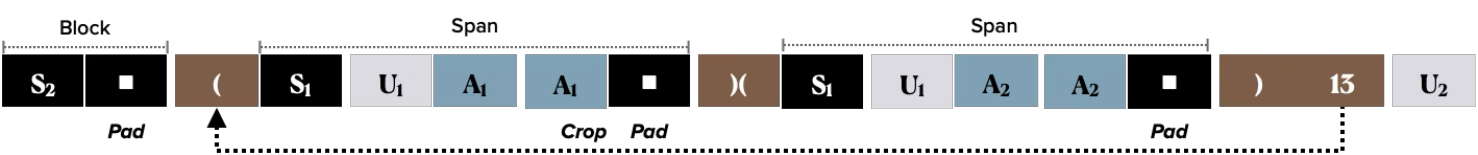}
        \caption{After low-level optimization pass}
    \end{subfigure}
    \caption{Tokenization and low-level optimization of the nested generation span query from \autoref{fig:span-query-nested-gen-optimized}. Block size of 2 tokens. Inner generates produce $A_1$ (3 tokens) and $A_2$ (2 tokens).
    }
    \label{fig:nested-gen-tokenization}
\end{figure}

This situation is visualized in \autoref{fig:nested-gen-tokenization}, which shows a tokenized span query for the nested generation example from \autoref{fig:span-query-nested-gen-optimized}. The output of the first inner generate is denoted $A_1$, which consists of three tokens in this example. Thus, the last token of the first inner generate will not be cached. Without deeper changes to vLLM (or to model fine tuning), we are forced to accept either cache misses (where, because of the dual output paradox, we send as input something different than was output to KV cache) or reduced accuracy (due to the need to crop, in order to prevent the paradox). 
For now, we accept that the third token of the first inner generate output $A_1$ must be cropped in order to get good cache locality.

To understand better why cache hit rate suffers and why padding alone cannot help, recall from \autoref{sec:background-vllm} that vLLM scans only for the \emph{prefix} of the blocks of a token sequence that are cache hits. Upon reaching a block that is not in cache, it scans no further for cache hits. If we do not crop the third token of $A_1$, vLLM's prefix scan will find a cache miss on the block that starts with the third token of $A_1$.

\subsection{Changes to the vLLM Scheduler Layer}
\label{sec:vllm-scheduler-changes}

Our first set of changes to vLLM is done to the ``scheduler'' layer of vLLM. This level of vLLM performs operations that are specific to a single client request. In particular, this logic includes KV cache management. In contrast, the layer below (\autoref{sec:vllm-gpu-runner-changes}) deals with optimizing GPU utilization across client requests.

When scanning a block sequence for cache hits, vLLM normally accumulates a hash code. This is how it computes a hash for each block in a way that depends on the hash of preceding blocks. We update this logic to allow disabling the incorporation of the accumulated hash into a block's hash. If the first token in a block is \texttt{(}, i.e. the start of a span sub-tree, we suspend the accumulation logic. If the first token in a block is \texttt{)}, i.e. the end of a span sub-tree, we resume the accumulation logic.

\subsection{Changes to the vLLM GPU Runner Layer}
\label{sec:vllm-gpu-runner-changes}

On a cache hit, the fetched block's KV vectors likely require updates before they can be reused. As visualized in \autoref{fig:chat-completion-illustration}, \autoref{fig:rag-illustration}, and \autoref{fig:nestedgen-illustration}, each cached KV vector has a positional encoding. If the position of a cached block (i.e. the position of its prior use) differs from its new position, adjustments are necessary. This adjustment is termed \emph{repositioning} --- a.k.a. a ReRoPE, which reverses and then re-applies a Rotational Positional Encoding~\cite{su2024roformer}.

In order to implement a high-throughput ReROPE we must tap into the lower ``GPU runner'' layer of vLLM. This will allow us to perform cross-request batching and to avoid race conditions between requests.

\subsubsection{CIDRA Repositioning Algorithm}
\label{sec:batch-repo-algo}

We implemented a ReROPE algorithm in the GPU runner layer.
Prior work~\cite{ma2025blockattention} does not handle the scenario where multiple concurrent requests reuse overlapping sets of KV vectors. To address this, we introduce the Concurrent In-place Duplicating ReROPE Algorithm (CIDRA). 


CIDRA first formulates the dependency graph of block repositionings --- e.g. block A moved to block B, B to C, and so on. 
This allows CIDRA to operate \emph{in-place}, minimizing scratch memory needed.
If a node has out degree greater than 1 (signaling that concurrent requests need to relocate a block but to different positions), CIDRA will duplicate that block. In other words, CIDRA duplicates nodes until it is left with a strict permutation graph.

The algorithm then performs a strongly connected components analysis of the repositioning graph. This analysis identifies the cycles and exposes the independent subgraphs that can be batched (via bin packing) for parallel GPU execution.
For large cycles exceeding the batch size, CIDRA falls back on CPU-based processing. This is rare in our experiments, where cycles are small (often size 2). 

\begin{figure}
    \centering
    \begin{subfigure}{0.47\linewidth}
        \includegraphics[width=\linewidth]{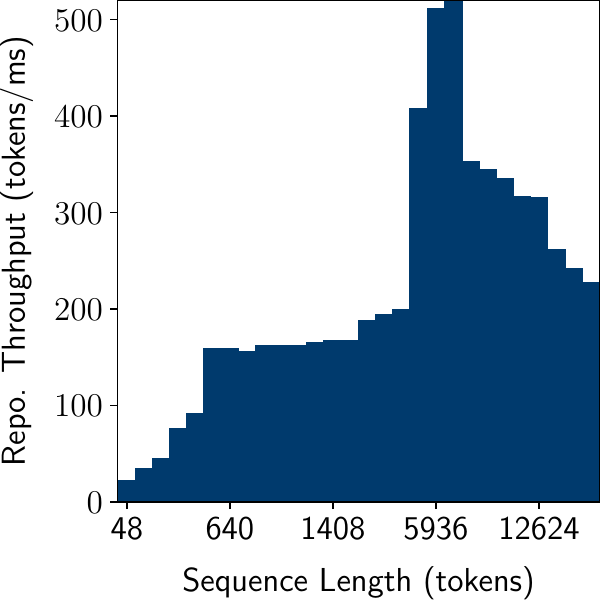}
        \caption{Repositioning throughput 
    }
        \label{fig:repo-throughput}
    \end{subfigure}
    \quad
    \begin{subfigure}{0.47\linewidth}
        \includegraphics[width=\linewidth]{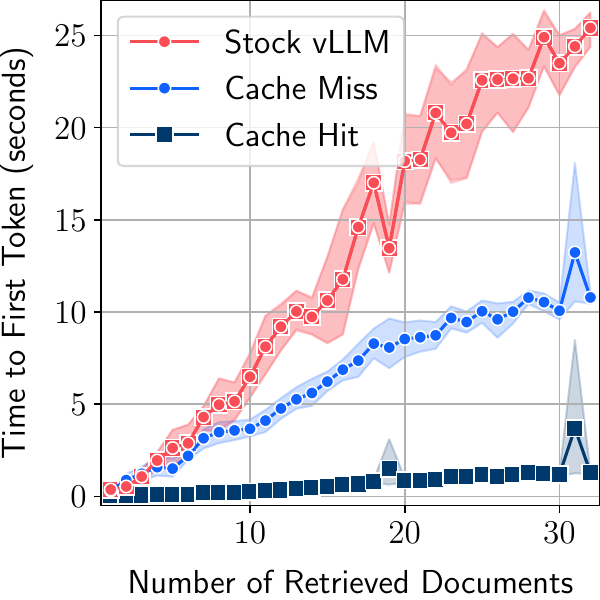}
      \caption{TTFT on RAG}
      \label{fig:rag-speedups}
    \end{subfigure}
    \caption{CIDRA repositioning throughput and TTFT of span query execution (using CIDRA) in a RAG microbenchmark. The right-hand figure compares stock vLLM execution to one using span queries. Even on cache miss, span queries are much faster.}
\end{figure}

Finally, CIDRA optimizes tensor operations by concatenating layers for small batches. 
\autoref{fig:repo-throughput} plots the repositioning throughput of CIDRA, showing a maximum throughput of 500 tokens per millisecond.

\subsection{TTFT Speedup on a RAG Span Query}
We present experimental results to quantify the effect our stack on KV cache locality. Our first RAG experiment uses a microbenchmark to probe the limits of what is possible with this use case.
In the microbenchmark, documents have 2857 tokens (the mean length of a Python source file in vLLM) of randomly generated content. The microbenchmark runs three variants across a range of 1--32 documents and measures TTFT for each experiment. Each tuple in this explored space is measured 10 times. The first (baseline) variant runs vLLM in its stock configuration. The documents will never be cache local in this configuration. The second variant measures the TTFT of our stack, but when the documents are not cache local. The third variant is the same, but in the case where the documents are served by cache. This variant will help us to understand our overheads.


We expect the TTFT of the first two variants to grow at worst quadratically as the number of documents increases. In the baseline (stock vLLM), every token in every document needs to attend to every prior token. For the second variant (span queries, cache miss), the presence of spans means that attention is sparse. The tokens within a document only attend to prior tokens in that same document. Both are quadratic at worst, but the second variant should have a smaller constant factor. Therefore, we expect a large performance benefit, even if we miss cache. The third variant (span queries, cache hit) should grow at worst linearly, as we can skip prefill but pay for repositioning. The experimental results, shown in \autoref{fig:rag-speedups}, bear this out.

\subsection{TTFT Speedup on a Nested Generate Span Query}

Next, we experiment with a nested generation microbenchmark. The goal is to compare TTFT for stock vLLM with the span query stack, for varying number of inner generates (this corresponds to the fan-out of the \SQPlus node) and for varying temperature of the inner generate. We should expect stock VLLM to perform well if the inner temperature is 0. In this case the output of the inner generates is the constant, and hence normal prefix caching works well. We should also expect stock vLLM to perform well if the fan-out is very small, because the cost of recomputation at some point becomes comparable to the cost of repositioning. 

\begin{figure}
    \centering
    \begin{subfigure}{0.495\linewidth}
        \includegraphics[width=0.9725\linewidth]{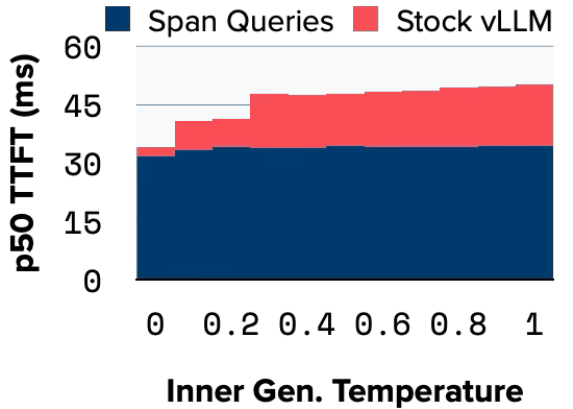}
        \caption{1 inner generate}
        \label{fig:nested-gen-ttft-speedups-varying-temperature-1}
    \end{subfigure}
    \begin{subfigure}{0.495\linewidth}
        \includegraphics[width=0.9725\linewidth]{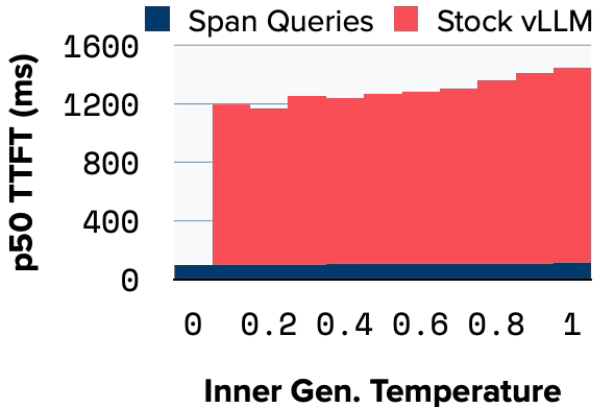}
        \caption{24 inner generates}
        \label{fig:nested-gen-ttft-speedups-varying-temperature-24}
    \end{subfigure}
    \caption{Nested generation microbenchmark, showing median TTFT and  varying inner generate temperature. 
    }
    \label{fig:nested-gen-ttft-speedups-varying-temperature}
\end{figure}

\autoref{fig:nested-gen-ttft-speedups-varying-temperature} corroborates this hypothesis. In this experiment, we executed a judge/generator span query 100 times with varying inner generate temperature. As shown, when inner temperature is 0, span query benefit to TTFT is near 0. For any non-zero inner temperature, the TTFT benefit depends on the inner fan-out. For an inner fan-out of 1 (\autoref{fig:nested-gen-ttft-speedups-varying-temperature-1}, the span query stack lowers TTFT by 7--32\% depending on inner temperature. For an inner fan-out of 24 (\autoref{fig:nested-gen-ttft-speedups-varying-temperature-24}), the span query stack has 12-13x faster TTFT.

\begin{figure}[ht]
    \centering

    \begin{subfigure}{0.495\linewidth}
          \includegraphics[width=0.9725\linewidth]{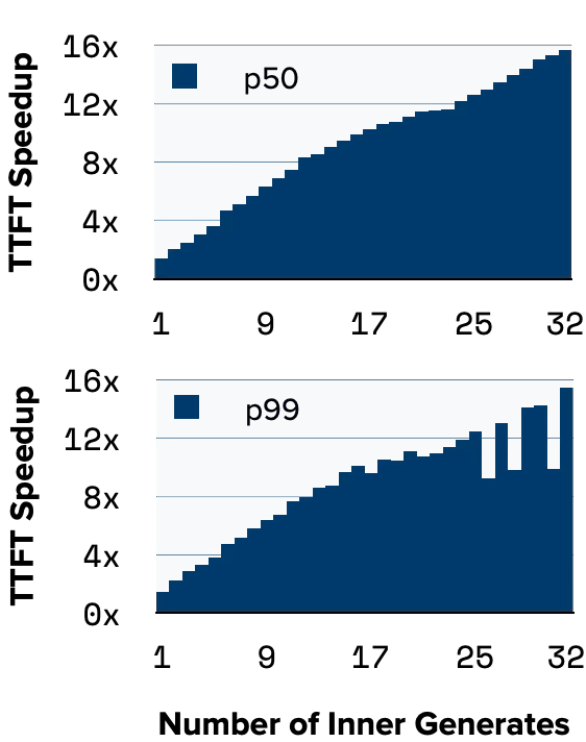}
    \caption{TTFT speedup}  
    \label{fig:nested-gen-ttft-speedups-temp-0.5-speedup}
    \end{subfigure}
        \begin{subfigure}{0.495\linewidth}
          \includegraphics[width=0.9725\linewidth]{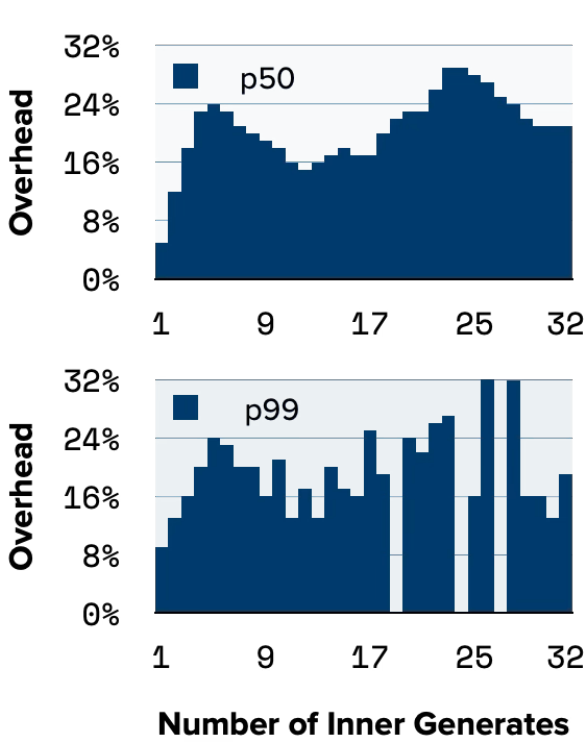}
    \caption{Repositioning overhead}  
    \label{fig:nested-gen-ttft-speedups-temp-0.5-overhead}
    \end{subfigure}

    \caption{Nested generation microbenchmark, showing the distribution of TTFT speedup from span queries and CIDRA overhead.
    }
    \label{fig:nested-gen-ttft-speedups-temp-0.5}
\end{figure}

\autoref{fig:nested-gen-ttft-speedups-temp-0.5-speedup} digs into this data to verify that the span query stack has stable performance. In this experiment, we used an inner temperature of 0.5, varied the inner fan-out, and ran 500 query executions for each valuoe of fan-out. We again compared stock vLLM to our stack. As shown, the TTFT speedup is stable up to the 99th percentile (p99), at which point stability begins to suffer. Further work is therefore necessary to ensure stability beyond two 9's. 

\autoref{fig:nested-gen-ttft-speedups-temp-0.5-overhead} shows the overhead of CIDRA repositioning. Observe the non-monotonic shape to the curves, and how these nicely trace the non-monotonic shape to the CIDRA repositioning throughput shown in \autoref{fig:repo-throughput}.

\subsection{TTFT Speedups from Bulk Span Query Execution}
\label{sec:results-accuracy}

\begin{table}
    \centering
    \caption{
    Bulk span query execution allows for scheduling the queries such that reused fragments are clustered temporally. Showing average speedup of TTFT by using span queries, for various bulk request sizes on two benchmarks.}
    \label{tab:results-bulk}
     \vskip 0.1in
    \begin{center}\begin{small}\begin{sc}
    \begin{tabular}{llll}
    \toprule
    Bulk Size & 2Wiki & NQ \\
    \midrule
    
         
        1 & 1.21x & 1.03x \\
        1024 & 1.31x & 1.05x \\
        whole corpus & 1.59x & 1.13x \\
         \bottomrule
    \end{tabular}
    \end{sc}\end{small}\end{center}
    \vskip -0.1in
\end{table}

We have explored the TTFT benefits of executing one span query at a time. We now consider an additional layer of potential offered by a bulk query execution API. If the queries in a bulk request are executed in the given order, KV cache locality may suffer if the working set of the bulk is larger than KV cache capacity. To experiment with this possibility, we implemented a greedy heuristic that clusters the requests in a given bulk to increase temporal locality. \autoref{tab:results-bulk} shows the average TTFT speedup (versus a stock vLLM baseline) of executing span queries in bulk. We explored two datasets:
2Wiki~\cite{ho-etal-2020-constructing} and NaturalQuestions~\cite{kwiatkowski-etal-2019-natural}
Both datasets are very small (each has around 100 tokens per fragment, placing them at the extreme left of \autoref{fig:rag-speedups}), hence as expected the speedup numbers align well with those presented in that figure.

\section{Improving Attention Locality}
\label{sec:locality-attention}

Finally, we show that the high-level optimizer from \autoref{sec:query-high-level-optimization} can be extended to target the ``lost in the middle'' problem~\cite{liu-etal-2024-lost}. In this well-studied phenomenon, model accuracy is U-shaped as a function of the position of relevant information within a given input. Models perform well if the ``needle`` is located near the beginning or end of the sequence, but poorly if it is buried in the middle of the ``hay''. We consider \emph{attention locality} as the success in finding the needle in the haystack. 

Prior work observes that one can increase attention locality by increasing the sparsity of the attention matrix~\cite{sun2025efficientattentionmechanismslarge}. We observe that a span query can be optimized to achieve the same effect. Unlike KV cache locality, where cache-optimized span queries rely on vLLM support, we now show how an attention-optimized span query can work without requiring any changes to the model server. 

\begin{figure}[ht]
    \centering
    \begin{subfigure}{\linewidth}
        \centering
        \includegraphics[width=0.75\linewidth]{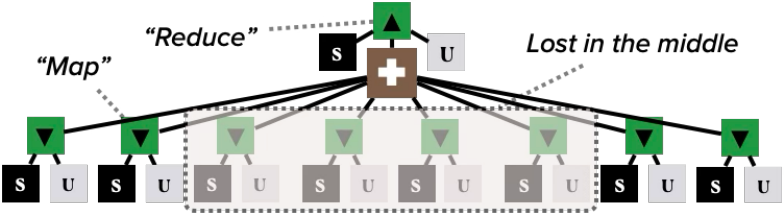}
        \caption{Unoptimized query, but with commutativity hint}
        \label{fig:attention-locality-query-unoptimized}
    \end{subfigure}
    \begin{subfigure}{\linewidth}
        \smallskip
        \centering
        \includegraphics[width=0.75\linewidth]{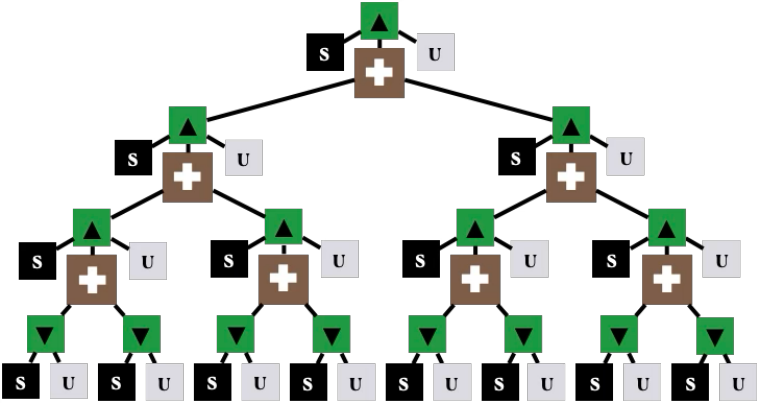}
        \caption{Optimized for attention locality}
        \label{fig:attention-locality-query-optimized}
    \end{subfigure}    
    \caption{High-level optimization for attention locality can transform a single 8-way judge/generator into 3 2-way judge steps. 
    }
    \label{fig:attention-locality-hlo}
\end{figure}

\autoref{fig:attention-locality-query-unoptimized} shows a judge/generator span query that generates 8 candidates and then judges them with an outer generate (c.f. \autoref{fig:nested-gen-query-with-plus-c}). The inner generates commute, as specified via \SQPlus. 
Alternatively, we could perform a tree reduction: e.g. judge 2 at a time, then judge the output of each pair of 2-way judgments, and so on through 3 plies. This query is visualized in \autoref{fig:attention-locality-query-optimized}. We have automated this by adding a query rewrite rule to the high-level optimizer.

\begin{figure}
    \centering

    \begin{subfigure}{0.495\linewidth}
        \centering
        \includegraphics[width=0.87\linewidth]{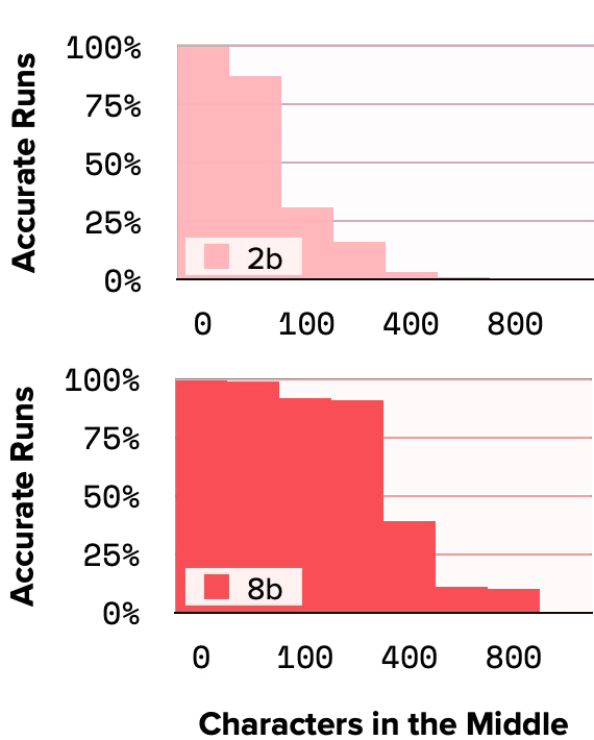}
        \caption{Unoptimized query}
    \end{subfigure}
    \begin{subfigure}{0.495\linewidth}
        \centering
        \includegraphics[width=0.87\linewidth]{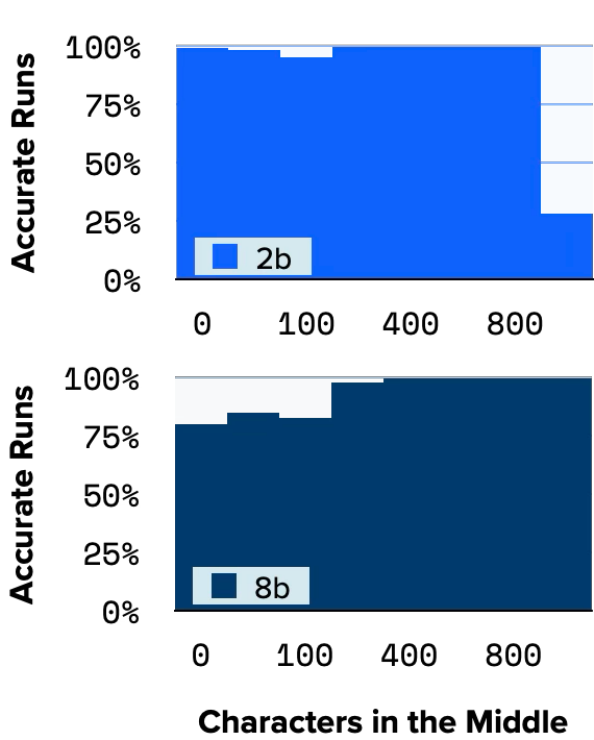}
        \caption{Attention-optimized $k=2$}
    \end{subfigure}
    \caption{Fraction of runs with good attention locality,
    as the length of the ``lost in the middle`` data increases. 
    }
    \label{fig:haystack}
\end{figure}

We implemented a simple needle-in-the-haystack microbenchmark.
The benchmark executes the two queries shown above, where the inner generates produce random names (the needles) interspersed with randomly generated content (the hay). The judge is tasked to extract the names from the hay. We run each query 1000 times, varying the amount of hay between the needles and using two variants of the granite3.3 model: 2b and 8b. Our hypothesis is that the unoptimized queries should perform well for smaller values of hay, and then drop off a cliff --- i.e. once the attention mechanism gets lost in the middle. Furthermore, the larger 8b model should tolerate more hay, as it is trained with a longer context length. And lastly, we hypothesize that the optimized queries should tolerate significantly higher amounts of hay. \autoref{fig:haystack} shows the results, which bear this out. Note how the 2b model, using the optimized query, fares better than the 8b model using the unoptimized query.
\section{Related Work}

Researchers are actively looking at techniques to improve cache performance, which is especially critical as cache size grows beyond GPU memory limits. Block Attention~\cite{ma2025blockattention} decomposes operations into discrete computational blocks, enhancing locality and enabling modular cache segmentation. CacheBlend~\cite{yao2025cacheblendfastlargelanguage} allows repositioning of fixed size cache blocks with greatly reduced re-calculation of attention between blocks while maintaining high accuracy. Superposition prompting improves RAG cache performance by breaking a prompt into parallel paths which allows for more selective attention. SGLang~\cite{zheng2024sglangefficientexecutionstructured} introduces a programming library to simplify LLM calling and allow more cache-aware optimizations. 
Research in Graphs-of-Thought~\cite{besta2024got} structures LLM processing into a graph structure that is inherently multi-generate and potentially allows for cache-aware placement of computation. Span Queries captures and extends these ideas into a declarative framework to co-optimize selective application of attention and cache locality.

\section{Conclusion}






We introduced the span query as an intermediate representation for analysis and optimization. Our focus to date has been on locality, for KV cache and for attention. There are many prospects for other optimization avenues. For example, is there any reason a nested generation structure should flow through KV cache at all? One could imagine analyzing a span query with nested generation to produce an optimized scheduler, where the inner and outer generates are arranged in a gather/scatter structure (c.f. SQL). This kind of optimization would benefit from span queries, because the query provides the full scope of the connectivity and ordering constraints --- the kind of information needed to make gather/scatter work well. This is one of many exciting avenues of exploration enabled by span queries.


\bibliography{spanqueries}

\begin{thebibliography}{21}
\providecommand{\natexlab}[1]{#1}
\providecommand{\url}[1]{\texttt{#1}}
\expandafter\ifx\csname urlstyle\endcsname\relax
  \providecommand{\doi}[1]{doi: #1}\else
  \providecommand{\doi}{doi: \begingroup \urlstyle{rm}\Url}\fi

\bibitem[Ayoub et~al.(2025)Ayoub, Harnik, Smith, Swain, Wang, Yin, and
  Yan]{llmd2025kvcache}
Ayoub, M., Harnik, D., Smith, T., Swain, K., Wang, X., Yin, H., and Yan, K.
\newblock From prefix caching in vllm to distributed scheduling with llm-d.
\newblock \url{https://llm-d.ai/blog/kvcache-wins-you-can-see}, September 2025.
\newblock URL \url{https://llm-d.ai/blog/kvcache-wins-you-can-see}.
\newblock Accessed: 2025-10-25.

\bibitem[Besta et~al.(2024)Besta, Blach, Kubicek, Gerstenberger, Gianinazzi,
  Gajda, Lehmann, Podstawski, Niewiadomski, Nyczyk, and Hoefler]{besta2024got}
Besta, M., Blach, N., Kubicek, A., Gerstenberger, R., Gianinazzi, L., Gajda,
  J., Lehmann, T., Podstawski, M., Niewiadomski, H., Nyczyk, P., and Hoefler,
  T.
\newblock {Graph of Thoughts: Solving Elaborate Problems with Large Language
  Models}.
\newblock \emph{Proceedings of the AAAI Conference on Artificial Intelligence},
  38\penalty0 (16):\penalty0 17682--17690, Mar 2024.
\newblock \doi{10.1609/aaai.v38i16.29720}.
\newblock URL \url{https://ojs.aaai.org/index.php/AAAI/article/view/29720}.

\bibitem[Davis et~al.(2024)Davis, Hanin, Chen, Bailis, Stoica, and
  Zaharia]{davis2024networksnetworkscomplexityclass}
Davis, J.~Q., Hanin, B., Chen, L., Bailis, P., Stoica, I., and Zaharia, M.
\newblock Networks of networks: Complexity class principles applied to compound
  ai systems design, 2024.
\newblock URL \url{https://arxiv.org/abs/2407.16831}.

\bibitem[Dean \& Ghemawat(2008)Dean and Ghemawat]{dean2008mapreduce}
Dean, J. and Ghemawat, S.
\newblock Mapreduce: simplified data processing on large clusters.
\newblock \emph{Communications of the ACM}, 51\penalty0 (1):\penalty0 107--113,
  2008.

\bibitem[Ho et~al.(2020)Ho, Duong~Nguyen, Sugawara, and
  Aizawa]{ho-etal-2020-constructing}
Ho, X., Duong~Nguyen, A.-K., Sugawara, S., and Aizawa, A.
\newblock Constructing a multi-hop {QA} dataset for comprehensive evaluation of
  reasoning steps.
\newblock In Scott, D., Bel, N., and Zong, C. (eds.), \emph{Proceedings of the
  28th International Conference on Computational Linguistics}, pp.\
  6609--6625, Barcelona, Spain (Online), December 2020. International Committee
  on Computational Linguistics.
\newblock \doi{10.18653/v1/2020.coling-main.580}.
\newblock URL \url{https://aclanthology.org/2020.coling-main.580/}.

\bibitem[Kwiatkowski et~al.(2019)Kwiatkowski, Palomaki, Redfield, Collins,
  Parikh, Alberti, Epstein, Polosukhin, Devlin, Lee, Toutanova, Jones, Kelcey,
  Chang, Dai, Uszkoreit, Le, and Petrov]{kwiatkowski-etal-2019-natural}
Kwiatkowski, T., Palomaki, J., Redfield, O., Collins, M., Parikh, A., Alberti,
  C., Epstein, D., Polosukhin, I., Devlin, J., Lee, K., Toutanova, K., Jones,
  L., Kelcey, M., Chang, M.-W., Dai, A.~M., Uszkoreit, J., Le, Q., and Petrov,
  S.
\newblock Natural questions: A benchmark for question answering research.
\newblock \emph{Transactions of the Association for Computational Linguistics},
  7:\penalty0 452--466, 2019.
\newblock \doi{10.1162/tacl_a_00276}.
\newblock URL \url{https://aclanthology.org/Q19-1026/}.

\bibitem[Kwon et~al.(2023)Kwon, Li, Zhuang, Sheng, Zheng, Yu, Gonzalez, Zhang,
  and Stoica]{10.1145/3600006.3613165}
Kwon, W., Li, Z., Zhuang, S., Sheng, Y., Zheng, L., Yu, C.~H., Gonzalez, J.,
  Zhang, H., and Stoica, I.
\newblock Efficient memory management for large language model serving with
  pagedattention.
\newblock In \emph{Proceedings of the 29th Symposium on Operating Systems
  Principles}, SOSP '23, pp.\  611–626, New York, NY, USA, 2023. Association
  for Computing Machinery.
\newblock ISBN 9798400702297.
\newblock \doi{10.1145/3600006.3613165}.
\newblock URL \url{https://doi.org/10.1145/3600006.3613165}.

\bibitem[Li et~al.(2025)Li, Li, Tian, Tang, Xu, Chen, Hu, Dong, Li, and
  Chen]{li2025surveylargelanguagemodel}
Li, H., Li, Y., Tian, A., Tang, T., Xu, Z., Chen, X., Hu, N., Dong, W., Li, Q.,
  and Chen, L.
\newblock A survey on large language model acceleration based on kv cache
  management, 2025.
\newblock URL \url{https://arxiv.org/abs/2412.19442}.

\bibitem[Liang et~al.(2024)Liang, Sun, Gui, Zhu, Jiang, Zhong, Qu, Zhao, Bo,
  Yang, Xiong, Yuan, Xu, Wang, Zhang, Zhang, Chen, Chen, and
  Zhou]{liang2024kagboostingllmsprofessional}
Liang, L., Sun, M., Gui, Z., Zhu, Z., Jiang, Z., Zhong, L., Qu, Y., Zhao, P.,
  Bo, Z., Yang, J., Xiong, H., Yuan, L., Xu, J., Wang, Z., Zhang, Z., Zhang,
  W., Chen, H., Chen, W., and Zhou, J.
\newblock Kag: Boosting llms in professional domains via knowledge augmented
  generation, 2024.
\newblock URL \url{https://arxiv.org/abs/2409.13731}.

\bibitem[Liu et~al.(2024)Liu, Lin, Hewitt, Paranjape, Bevilacqua, Petroni, and
  Liang]{liu-etal-2024-lost}
Liu, N.~F., Lin, K., Hewitt, J., Paranjape, A., Bevilacqua, M., Petroni, F.,
  and Liang, P.
\newblock Lost in the middle: How language models use long contexts.
\newblock \emph{Transactions of the Association for Computational Linguistics},
  12:\penalty0 157--173, 2024.
\newblock \doi{10.1162/tacl_a_00638}.
\newblock URL \url{https://aclanthology.org/2024.tacl-1.9/}.

\bibitem[Ma et~al.(2025)Ma, Wang, and Lan]{ma2025blockattention}
Ma, D., Wang, Y., and Lan, T.
\newblock Block-attention for efficient prefilling.
\newblock In \emph{The Thirteenth International Conference on Learning
  Representations}, 2025.
\newblock URL \url{https://openreview.net/forum?id=7zNYY1E2fq}.

\bibitem[Merth et~al.(2024)Merth, Fu, Rastegari, and
  Najibi]{merth2024superpositionpromptingimprovingaccelerating}
Merth, T., Fu, Q., Rastegari, M., and Najibi, M.
\newblock Superposition prompting: Improving and accelerating
  retrieval-augmented generation, 2024.
\newblock URL \url{https://arxiv.org/abs/2404.06910}.

\bibitem[{PwC}(2025)]{PwC_AIAgentSurvey_2025}
{PwC}.
\newblock Pwc’s ai agent survey.
\newblock Tech Effect: AI \& Analytics, PwC US, May 2025.
\newblock URL
  \url{https://www.pwc.com/us/en/tech-effect/ai-analytics/ai-agent-survey.html}.
\newblock Survey of 308 US executives conducted Apr 22--28, 2025.

\bibitem[Sarthi et~al.(2024)Sarthi, Abdullah, Tuli, Khanna, Goldie, and
  Manning]{sarthi2024raptorrecursiveabstractiveprocessing}
Sarthi, P., Abdullah, S., Tuli, A., Khanna, S., Goldie, A., and Manning, C.~D.
\newblock Raptor: Recursive abstractive processing for tree-organized
  retrieval, 2024.
\newblock URL \url{https://arxiv.org/abs/2401.18059}.

\bibitem[Snell et~al.(2025)Snell, Lee, Xu, and Kumar]{snell2025scaling}
Snell, C.~V., Lee, J., Xu, K., and Kumar, A.
\newblock Scaling {LLM} test-time compute optimally can be more effective than
  scaling parameters for reasoning.
\newblock In \emph{The Thirteenth International Conference on Learning
  Representations}, 2025.
\newblock URL \url{https://openreview.net/forum?id=4FWAwZtd2n}.

\bibitem[Su et~al.(2024)Su, Ahmed, Lu, Pan, Bo, and Liu]{su2024roformer}
Su, J., Ahmed, M., Lu, Y., Pan, S., Bo, W., and Liu, Y.
\newblock Roformer: Enhanced transformer with rotary position embedding.
\newblock \emph{Neurocomputing}, 568:\penalty0 127063, 2024.

\bibitem[Sun et~al.(2025)Sun, Li, Zhang, Pan, Dong, Guo, and
  Wang]{sun2025efficientattentionmechanismslarge}
Sun, Y., Li, Z., Zhang, Y., Pan, T., Dong, B., Guo, Y., and Wang, J.
\newblock Efficient attention mechanisms for large language models: A survey,
  2025.
\newblock URL \url{https://arxiv.org/abs/2507.19595}.

\bibitem[Vaswani et~al.(2017)Vaswani, Shazeer, Parmar, Uszkoreit, Jones, Gomez,
  Kaiser, and Polosukhin]{vaswani2017attention}
Vaswani, A., Shazeer, N., Parmar, N., Uszkoreit, J., Jones, L., Gomez, A.~N.,
  Kaiser, {\L}., and Polosukhin, I.
\newblock Attention is all you need.
\newblock \emph{Advances in neural information processing systems}, 30, 2017.

\bibitem[Yao et~al.(2025)Yao, Li, Liu, Ray, Cheng, Zhang, Du, Lu, and
  Jiang]{yao2025cacheblendfastlargelanguage}
Yao, J., Li, H., Liu, Y., Ray, S., Cheng, Y., Zhang, Q., Du, K., Lu, S., and
  Jiang, J.
\newblock Cacheblend: Fast large language model serving for rag with cached
  knowledge fusion, 2025.
\newblock URL \url{https://arxiv.org/abs/2405.16444}.

\bibitem[Zheng et~al.(2024)Zheng, Yin, Xie, Sun, Huang, Yu, Cao, Kozyrakis,
  Stoica, Gonzalez, Barrett, and
  Sheng]{zheng2024sglangefficientexecutionstructured}
Zheng, L., Yin, L., Xie, Z., Sun, C., Huang, J., Yu, C.~H., Cao, S., Kozyrakis,
  C., Stoica, I., Gonzalez, J.~E., Barrett, C., and Sheng, Y.
\newblock Sglang: Efficient execution of structured language model programs,
  2024.
\newblock URL \url{https://arxiv.org/abs/2312.07104}.

\bibitem[Zhou et~al.(2025)Zhou, Xu, Wang, Xiong, and
  Joty]{zhou2025evaluatingjudgesevaluatorsjetts}
Zhou, Y., Xu, A., Wang, P., Xiong, C., and Joty, S.
\newblock Evaluating judges as evaluators: The jetts benchmark of llm-as-judges
  as test-time scaling evaluators, 2025.
\newblock URL \url{https://arxiv.org/abs/2504.15253}.

\end{thebibliography}
\bibliographystyle{mlsys2025}

\end{document}